\documentclass{article}

\usepackage{microtype}
\usepackage{graphicx}
\usepackage{subfigure}
\usepackage{booktabs} 

\usepackage{amsmath}
\usepackage{etoolbox}

\usepackage{amssymb}
\usepackage{algorithmic}
\usepackage{forloop}
\usepackage{fancyhdr}
\usepackage{stmaryrd}
\usepackage{textcomp}
\usepackage{bm}

\usepackage{amssymb}
\usepackage{booktabs}
\usepackage{multirow}
\usepackage{kotex}
\usepackage{tablefootnote}
\usepackage{footnote}
\usepackage[dvipsnames,table,xcdraw]{xcolor}
\usepackage{lineno}
\usepackage{hyperref}
\DeclareMathOperator*{\argmax}{arg\,max}

\usepackage{hyperref}

\usepackage[accepted]{icml2021}

\icmltitlerunning{K-Wav2vec 2.0: Automatic Speech Recognition based on Joint Decoding of Graphemes and Syllables}

\begin{document}

\twocolumn[
\icmltitle{K-Wav2vec 2.0: Automatic Speech Recognition based on Joint Decoding of Graphemes and Syllables}



\icmlsetsymbol{equal}{*}

\begin{icmlauthorlist}
\icmlauthor{Jounghee Kim}{Korea}
\icmlauthor{Pilsung Kang}{Korea}
\end{icmlauthorlist}

\icmlaffiliation{Korea}{School of Industrial Management Engineering,
College of Engineering,
Korea University, Seoul, Korea}

\icmlcorrespondingauthor{Pilsung Kang}{pilsung\_kang@korea.ac.kr}

\icmlkeywords{Cross-lingual transfer, Joint decoding, Multitask learning, Speech recognition, Wav2vec 2.0}

\vskip 0.3in
]



\printAffiliationsAndNotice{ } 

\begin{abstract}
Wav2vec 2.0 is an end-to-end framework of self-supervised learning for speech representation that is successful in automatic speech recognition (ASR), but most of the work on the topic has been developed with a single language: English. Therefore, it is unclear whether the self-supervised framework is effective in recognizing other languages with different writing systems, such as Korean which uses the Hangul having a unique writing system. In this paper, we present K-Wav2Vec 2.0, which is a modified version of Wav2vec 2.0 designed for Korean automatic speech recognition by exploring and optimizing various factors of the original Wav2vec 2.0. In fine-tuning, we propose a multi-task hierarchical architecture to reflect the Korean writing structure. Moreover, a joint decoder is applied to alleviate the problem of words existing outside of the vocabulary. In pre-training, we attempted the cross-lingual transfer of the pre-trained model by further pre-training the English Wav2vec 2.0 on a Korean dataset, considering limited resources. Our experimental results demonstrate that the proposed method yields the best performance on both Korean ASR datasets: Ksponspeech (a large-scale Korean speech corpus) and Clovacall (a call-based dialog corpus). Further pre-training is also effective in language adaptation, leading to large improvements without additional data.
\end{abstract}

\section{Introduction}
\label{sec:intro}

In recent years, self-supervised methodology, wherein a model is pre-trained with abundant unlabeled data and subsequently fine-tuned with more restricted labeled data, has shown success in various fields, including natural language processing \cite{devlin2019bert, liu2019roberta}, image recognition \cite{chen2020simple}, and auto speech recognition  \cite{schneider2019wav2vec, baevski2019vq, baevski2020wav2vec}. The Wav2vec 2.0 model \cite{baevski2020wav2vec} is an end-to-end framework of self-supervised learning for automatic speech recognition (ASR), and it has recently been presented as an effective pre-training method to learn speech representations. When followed by fine-tuning with small amounts of labeled data, the Wav2vec 2.0 model has shown remarkable performance in English ASR tasks. However, despite the model's great success, it is still an open question whether this method can be effective with other languages, because all experiments conducted to date have been only with English datasets such as Librispeech \cite{panayotov2015librispeech} and TIMIT \cite{garofolo1993timit}.
In this paper, we introduce a novel adaptation of the Wav2vec 2.0 model that allows it to handle Korean ASR by considering various language-specific features, an effective fine-tuning architecture, an efficient pre-training method, and appropriate datasets for pre-training.

In the Korean writing system, letters are written in syllabic blocks, where one sound is made at once. These syllabic blocks are composed of 51 Korean grapheme units, including 30 consonants and 21 vowels. This unique writing system allows us to build a Korean ASR model that is based on either graphemes or syllable blocks. So far, most Korean ASR models  \cite{kim2020kospeech, bang2020ksponspeech, ha2020clovacall} were developed with a high-level modeling unit, syllables, and only a few studies attempted to use graphemes \cite{park2019korean, lee2019korean}. According to previous research that investigated modeling units in Korean ASR tasks, syllable-based models outperform grapheme-based models on Zeroth-Korean dataset (51.6 hrs) in most cases \cite{wang2020exploring}. Because they require more combinations to predict, grapheme-based models generally underperform relative to syllable-based models. However, syllable-based models also have data sparseness problem for infrequently used syllables and the out-of-vocabulary (OOV) problem when the training data is insufficient. In other languages which suffer from the same issues, such as English, Mandarin Chinese, and Japanese, previous research has identified methods \cite{sanabria2018hierarchical, krishna2018hierarchical, chen2021investigation, ueno2018acoustic} to overcome these problems using a multi-task learning approach (MTL). Multi-task learning is a method for learning shared representations from different (but related) tasks using different modeling units together. By learning the shared representations between high-level and low-level modeling units, multi-task models alleviate data sparseness issues and achieve better performance  \cite{chen2021investigation}.
Moreover, some research \cite{ueno2018acoustic} has introduced recovering methods to substitute OOV words produced for high-level decoding with segments generated in low-level outputs.

Inspired by prior works conducted with other languages, we propose a multi-task hierarchical fine-tuning architecture of Wav2vec 2.0 to reflect the unique relationship that exists in Korean writing between syllables and graphemes. By learning useful intermediate representations, the proposed model can generate multi-level units without sacrificing performance. In the inference step, we used a joint decoding strategy that considers high-level and low-level units to find the best sequence from a set of candidates instead of using additional language models. This decoding approach leads to better performance and alleviates the problem of OOV words in high-level outputs.

In practice, although unlabeled data is much easier to obtain than labeled data, collecting appropriate unlabeled data for stable training is still expensive. To consider the practical circumstance wherein only limited resources are available, we also experimented with the cross-lingual transfer of the English Wav2vec 2.0 model to Korean ASR tasks. Recently, several works \cite{riviere2020unsupervised, conneau2020unsupervised} have shown that the cross-lingual transfer methods, which include pre-training with English and fine-tuning with the other low-resource languages, can be effective in improving the performance of downstream tasks \cite{riviere2020unsupervised}. We adopted the cross-lingual transfer method to improve the model’s performance with a limited Korean dataset by further pre-training the English Wav2vec 2.0 model.
More specifically, we pre-train the English Wav2vec 2.0 model on a Korean dataset first and fine-tuned the model later. Our further pre-training efficiently learns Korean speech representations, taking advantage of learned representations from another language, and it achieves better performance than the pre-trained model from scratch.

The remainder of this paper is organized as follows. Section \ref{sec:backgrounds} briefly reviews the training process for the Wav2vec 2.0. Section \ref{sec:method} presents the proposed method, including the fine-tuning architecture, joint decoder, and pre-training method. Section \ref{sec:experimental_settings} introduces the details of the experimental configuration, and Section \ref{sec:experimental_results} shows the experimental results, verifying the effectiveness of our method. Section \ref{sec:discussions} discusses the various properties of the proposed method that improve performance. Finally, we conclude our study in Section \ref{sec:conclusion}.

\section{Background}
\label{sec:backgrounds}

In this section, we briefly review the base architecture of Wav2vec 2.0 using the pre-training method. Then, we explain the fine-tuning process using a pre-trained network for ASR.

\subsection{Pre-training Wav2vec 2.0}
The base architecture of Wav2vec 2.0 consists of three networks: a feature encoder, a contextual transformer, and a quantization module \cite{baevski2020wav2vec}. The feature encoder, which is composed of a multi-layer convolutional neural network (CNN), encodes the raw audio $X$ and outputs the latent speech representations $Z$. The contextual transformer, which is a stack of transformer encoders, learns the context representation $C$ by taking latent speech representations as input. The quantization module is used to map latent representations into the discretized space $Q$, choosing discrete codebook entries in a fully differentiable way.

In pre-training, a certain portion of the latent representations are randomly masked before feeding them into the contextual transformer. The model is trained by solving a contrastive task with masked representations, distinguishing the true quantized latent vector from those discrete latent vectors that have been randomly sampled from the other masked time steps. During pre-training, the model learns contextualized representations from only the unlabeled speech audio data.

\subsection{Fine-tuning for ASR}

For ASR tasks, a randomly initialized linear layer is added on top of the pre-trained model. This linear layer takes the contextualized representations of the pre-trained model and generates the most probable words. In fine-tuning, connectionist temporal classification (CTC) loss, which is an approach for sequence labeling without requiring alignment information between the output sequences and the input audio, is used to train both the linear layer and the pre-trained model.

\begin{figure}[t!]
\vskip 0.2in
\begin{center}
\centerline{\includegraphics[width=\columnwidth]{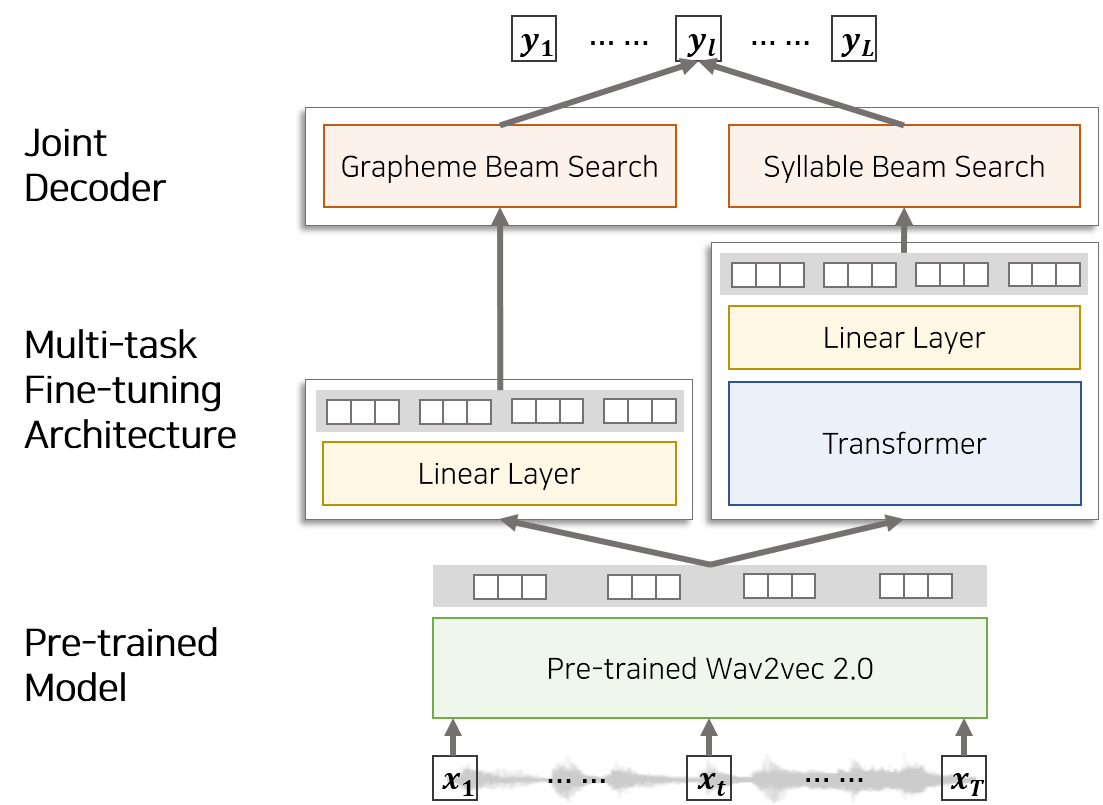}}
\caption{Overview of our proposed ASR framework. The pre-trained model encodes raw audio into feature vectors. The multi-task architecture simultaneously predicts probabilities of both grapheme and syllable-level sequences using the encoded audio features. The joint decoder aggregates these two outputs to determine the most confident sequence.}
\label{fig:overview}
\end{center}
\vskip -0.2in
\end{figure}

\section{Method}
\label{sec:method}

Figure \ref{fig:overview} shows the architecture of our Korean ASR system. The proposed system is a stack of the pre-trained Wav2vec 2.0, the multi-task fine-tuning architecture, and the joint decoder. 
The second and third components of the proposed model are described in more detail in the following sections.

\subsection{Multi-task hierarchical architecture}

The multi-task fine-tuning architecture, which consists of the grapheme encoder and the syllable encoder, is trained by taking contextualized representations learned from the pre-trained Wav2vec 2.0 model as an input. Given the raw audio $X$, let $C=c_1, ..., c_F$ be the sequence of encoded audio features, that is, the contextualized representations of the pre-trained model, where $F \in \mathbb{N}^+$ is the number of encoded audio frames. In the grapheme encoder, a linear layer is adopted to project the encoded features into a grapheme vocabulary $g \in G$, followed by the softmax function to produce the posterior probabilities of the grapheme sequences $p(g_f|c_f)$ corresponding to each frame. In the syllable encoder, alternatively, the encoded features are first fed to a stack of transformer encoders, which converts the encoded features $c_f$ to a sequence of hidden vectors $h_f$ to capture the relationship between low-level and high-level information. Then, a linear projection layer and the softmax function are applied to produce the posterior probabilities of the syllable sequences $p(s_f|h_f)$, where $s \in S$ is the syllable vocabulary. Under the conditional independence assumption, the posterior probability of a complete sequence is computed using the syllable outputs or grapheme outputs of each frame:
\begin{gather*}					
p_{syll.}(l=s_1,...s_F|X)=\prod_{f=1}^{F}{p(s_f|h_f)}, \label{p-syllable} \\
p_{grap.}(l=g_1,...g_F|X)=\prod_{f=1}^{F}{p(g_f|c_f)}.
\label{p-grapheme}
\end{gather*}

\subsection{Multi-task fine-tuning}

We used the CTC loss when training the model \cite{graves2006connectionist}. The CTC is an approach for sequence labeling, wherein the lengths of the label sequence and the output frames are different. Given a ground truth label sequence $Y=y_1, ... y_l$, the CTC process can expand $Y$ to a set of sequences $\Omega(Y)$ by adding a blank token between consecutive labels and allowing each label to be repeated, where the length of each sequence in the extended set $\Omega(Y)$ is the same as the length of the output frames. Then the posterior conditional probability of label sequence with the CTC process is computed as follows:
\begin{gather*}
p^{ctc}_{syll.}(Y|X)=\sum_{l \in \Omega(Y)}{p_{syll.}(l | X)}=\sum_{l \in \Omega(Y)}{\prod_{f=1}^{F}{p(s_f|h_f)}}.
\label{ctc-posterior}
\end{gather*}
The CTC loss is the log posterior conditional probability of the label sequence with the CTC process and it can be efficiently computed by using the forward-backward algorithm. The objective function of the proposed architecture is the weighted sum of grapheme and syllable-level CTC losses to learn the two tasks simultaneously:
\begin{gather*}
L_{MTL} = \lambda \log p^{ctc}_{syll.} (Y|X) + (1-\lambda) \log p^{ctc}_{grap.} (Y|X), \label{L-MTL}
\end{gather*}
where $\lambda:0 \leq \lambda \leq 1$ is an adjustable hyper-parameter that controls the trade-off between the importance of syllables and graphemes.
By setting an appropriate $\lambda$, our proposed model can learn the intermediate relation between the syllable and graphene by minimizing the multi-task loss.

\subsection{Joint decoder}

During the inference step, decoding algorithms are generally used to identify confident candidates. Because inference time is limited in practice, searching all possible combinations (i.e., brute-force approach) is computationally prohibitive. Instead, heuristic local search algorithms, such as greedy search, Viterbi, and beam search, are commonly used. 
In our proposed system, the joint decoder combines low-level and high-level beam search results to find the best sequence within a limited amount of time. In contrast to prior works \cite{hori2017multi, hori2017joint}, we only used outputs of the acoustic model, which allowed us to build ASR system without requiring external language models.

To extract the candidate sequences from the outputs with the CTC process, we used the CTC beam search decoder described in \cite{hannun2017sequence}. The beam search decodes iteratively to find candidates over time-steps of CTC output and scores them with given probabilities of each time-step. Given a set of syllable outputs, the beam search decoding returns several of the most likely sequences, syllable candidates $\hat{S}$, and their corresponding probabilities. The beam search results of the grapheme outputs are grapheme candidates $\hat{G}$ and their probabilities in the same way. The objective of the joint decoder is to find the most probable sequence $\hat{Y}$ among the candidates:
\begin{gather*}
\hat{Y}=\argmax_{Y \in \hat{S} \cup \hat{G}} \{ \gamma p^{ctc}_{syll.}(Y|X) + (1-\gamma) p^{ctc}_{grap .} (Y|X) \}, \label{ joint-decoder objectives }
\end{gather*}
where $\gamma:0 \leq \gamma \leq 1$ is a weight that controls the contribution of the results from two different levels to the final output. In the joint decoding process, the proposed model can be more robust than a single encoder (i.e., the syllable-based decoder or the grapheme-based decoder) by adjusting the probability of the syllable and grapheme candidates when the same candidate is in each beam search result $\hat{S} \cap \hat{G}$. The additional candidates for syllable output can be generated using the grapheme candidates, where non-overlapped candidates of the grapheme beam search exist $\hat{G} - (\hat{S} \cap \hat{G})$.
Because the grapheme modeling unit has less vocabulary but can express more syllables by combining graphemes, the additional candidates from the grapheme beam search result alleviate the OOV problem and the data sparseness problem pertaining to syllable output.

\subsection{Cross-lingual transfer pre-training}

In this section, we explain our pre-training approach for cross-lingual transfer of the English model to Korean. In practice, the collection of labeled data not only requires significant resources, but it also needs high-quality unlabeled data that affect training stability, which are expensive to obtain. To leverage the ASR performance with limited data, we explore cross-lingual transfer by further pre-training the existing English model with a Korean dataset. Many studies \cite{riviere2020unsupervised, conneau2020unsupervised, bansal2019pre} have investigated the cross-lingual transfer of speech representations by pre-training the model using a high-resource language and fine-tuning to the lower-resource ones. Although a pre-trained model from a different language reportedly enhances performance, it is not specialized to adapt the target languages. Hence, we propose further pre-training on the target language data (i.e., the Korean dataset) after pre-training on a high resource language (i.e., English).

The motivation behind this approach was the early successes of further pre-training in natural language processing approaches \cite{gururangan2020don, sun2019fine, beltagy2019scibert, lee2020biobert}. Previous researchers investigated the impact of further pre-training with various language domains by pre-training the models on a task-relevant corpus, followed by a classification task of the target domains. Recent research \cite{gururangan2020don} has demonstrated the benefit of further pre-training the generalized models, such as BERT \cite{devlin2019bert} and RoBERTa \cite{liu2019roberta}, on target-specific data to specialize the models in the domain of their target task. Insofar as the speech representations of the pre-trained models are known to share generalized features across languages \cite{riviere2020unsupervised}, further pre-training on the target language can also benefit by taking advantage of this shared information.
In the proposed model, we further pre-train the English Wav2vec 2.0, which was originally pre-trained on the English dataset (960 h), on the Korean dataset (965 h). When this additional pre-training is completed, the model is fine-tuned with Korean-labeled speech data for the downstream task.

\section{Experimental Settings}
\label{sec:experimental_settings}

\subsection{Datasets}
\label{subsection:datasets}

\begin{table}[t!]
    \centering
    \begin{tabular}{lcccc}
    \toprule 
         \multirow{2}{*}{\small Transcription} & \multirow{2}{*}{Unit} & \multirow{2}{*}{\small \# Vocabs.} & \multicolumn{2}{c}{\small \# OOV} \\ 
         & & & \small clean & \small other \\ \hline
        
        \multirow{2}{*}{\small Clovacall-Base} & \footnotesize syllable & 625 & 40 & - \\
        & \footnotesize grapheme & 82 & 7 & - \\ \hline
        
        \multirow{2}{6em}{\small Ksponspeech \newline \textit{\small - phonetic}} & \footnotesize syllable & 2260 & 3 & 3 \\
        & \footnotesize grapheme & 67 & 0 & 0 \\ \hline
        
        \multirow{2}{6em}{\small Ksponspeech \newline \textit{\small - orthographic}} & \footnotesize syllable & 2302 & 1 & 2 \\
        & \footnotesize grapheme & 104 & 0 & 0 \\
          
    \bottomrule
    \end{tabular}{\parfillskip=0pt\par}
    \caption{OOV with each modeling unit in the evaluation sets (evaluation-clean/evaluation-other). \# Vocabs refers to the number of vocabularies in the training set. The Ksponspeech provides a dual transcription, converted into a phonetic and orthographic script with appropriate preprocessing methods.}
    \label{tab:oov_information}
\end{table}

We verify the proposed method with Korean speech datasets, including a large-corpus dialog dataset, Ksponspeech \cite{bang2020ksponspeech}, and a call-based benchmark speech corpus, Clovacall \cite{ha2020clovacall}.

Ksponspeech is an open-domain dialog corpus recorded with 2,000 native Korean speakers in a controlled, quiet environment setting. The standard split datasets consist of 965 h of training, 4 h of development, 3 h of evaluation-clean, and 4 h of evaluation-other. We followed these data splits to train and evaluate the models. 
The dual transcription was provided for downstream tasks: one was a transcription with Korean standard orthographic rules and with the other retained the original sound.
The orthographic transcription contains numerical notation and abbreviation notation, such as ``SOTA,’’ meaning ``state-of-the-art.’’ However, phonetic transcription consists of only Korean characters that convert numerical symbols and abbreviations to the original sound. 
Following the preprocessing guidelines outlined in previous research \cite{kim2020kospeech}, we split the dual transcription into orthographic and phonetic scripts to evaluate our method.

Clovacall is a customer-service dialog corpus recorded by 11,000 people via phone calls. It has relatively short dialogs, most of which are goal-oriented, such as reservation or delivery. We only used publicly opened parts of the dataset, known as Clovacall-Base. We used the evaluation splits of Clovacall, comprising 50 h of labeled data for training and 1 h of data for testing. We randomly sampled 10\% of the training set for use as a development set for validation. The amount of Clovacall training data was relatively small to cover all the vocabulary of the evaluation set. Table \ref{tab:oov_information} shows that there are more OOVs in Clovacall than in Ksponspeech for both modeling units. Moreover, a data sparseness problem exists in Clovacall, wherein the occurrence of less common syllables is very low.

\subsection{Pre-trained model}

To build the Korean adapted model (i.e., K-Wav2vec 2.0), we utilized the English Wav2vec 2.0 released by \cite{baevski2020wav2vec}, which is pre-trained on 960 h of Librispeech without fine-tuning. We further pre-trained the English model on 965 h of Ksponspeech data for 400k updates. In our experiments, we mostly followed the experimental settings of the English model to pre-train the models, including a learning rate of 5e-4, an Adam optimizer \cite{kingma2015adam}, and 32k warm-up steps. For stable pre-training, we used silence elimination similar to that in \cite{kim2020kospeech}, excluding any ranges under 30 dB in raw audio. Because the dataset was recorded in a quiet environment, eliminating prolonged silence in conversation makes the model converge fast and is memory efficient. We also cropped audio samples longer than 15.6 s or shorter than 1 s and used samples having a length of 87.5 s as a batch per GPU. To meet the standard pre-training setting, we used two V100 GPUs with 32 gradient accumulations for each update. We chose the checkpoint having the lowest contrastive loss in the development subset.

\subsection{Fine-tuning strategies}
\label{subsection:fine_tune_strategies}

After pre-training, we conducted multi-task fine-tuning, as shown in Figure \ref{fig:overview}. The graphene encoder is a single linear layer, and the syllable encoder is a stack of 2 transformer blocks, which contains model dimension 768 and 8 attention heads, and a linear layer. These two modules were fine-tuned using labeled training data with multi-task loss. We used a hyper-parameter, $\lambda$, of 0.5 for multi-task loss to reflect the grapheme and syllable CTC loss equally. We trained models with the Adam optimizer and a tri-state rate scheduler, where a learning rate of 1e-5 was applied to Ksponspeech and 3e-5 on Clovacall. In the low-resource experiment using Clovacall, we froze the parameters of pre-trained parts for the first 10k updates to train the multi-task architecture first. Then, all parameters, including the pre-trained network and the fine-tuning architecture, were optimized for 70k updates. With large-corpus data, we fine-tuned all parameters together for 320k updates without an initial freezing state. The modified SpecAugment \cite{baevski2020wav2vec}, a strategy to make the model robust to noise by randomly masking embedding spans, was also used. For evaluation, we chose the checkpoint having the lowest word error rate (WER) on the development subset. To make the decoding process computationally efficient in the training phase, we selected the maximum probability syllables of each sequence and decoded it based on the inverse CTC process to collapse consecutive labels and delete the blank tokens. More details are available online\footnote{\url{https://github.com/JoungheeKim/K-wav2vec}}.

\subsection{Evaluation metrics}

We used the character error rate (CER), WER, and space-normalized WER (sWER), all of which are commonly used in Korean ASR as performance evaluation metrics. Note that the grapheme outputs are converted to syllables before evaluation for a fair comparison. The CER is calculated at the syllable level using the Levenshtein distance between the actual sequence of transcription and predicted sequence of the model. The WER is derived from the Levenshtein distance with the word level. The sWER is a modified version of the word error rate used to evaluate Korean ASR, wherein space rules are flexible \cite{bang2020ksponspeech}. In Korean datasets, speech transcriptions have sequences of inconsistent spacing, which results in invalid evaluations. Therefore, pseudocode was used to modify the predicted sequences by normalizing spaces based on actual sequences, and the sWER was calculated with these space-normalized sequences. 

\subsection{Baseline}

We employed transformer-based ASR models consisting of transformer blocks and a VGG encoder released in ESPnet as a baseline model \cite{watanabe2018espnet}. Following the structures described in \cite{bang2020ksponspeech}, we reproduced the large transformers for the large-corpus dataset, in which the multi-head attention is composed of eight attention heads with 512 hidden dimensions. Considering the small vocabulary, we used small transformers consisting of four multi-head attention with 256 hidden dimensions for Clovacall. Syllable is the unit of the baseline model, and we used the same beam search algorithm with 60 beam sizes as in \cite{bang2020ksponspeech}.

\section{Experimental Results}
\label{sec:experimental_results}

To investigate the effect of multitask fine-tuning and decoding strategies, we built six K-Wav2vec 2.0, as shown in Table \ref{tab:high_resource}. The architecture in the parentheses in the first column (model) is the fine-tuning structure, and the second column denotes the decoding scheme used. The last row of each transcription, K-Wav2vec 2.0, is our proposed model, and the other five models lack some parts of the proposed model. The transformer for fine-tuning architecture is identical to the syllable encoder of the multi-task model, and the linear is the same as the grapheme encoder. For all models, we used a beam size of 100 for the decoding process, and a $\gamma$ of 0.5 was applied to the joint decoder.

\begin{table*}[t]
    \centering
    \begin{tabular}{lcccccccccc}
    \toprule 
         \multirow{2}{*}{Model} & \multirow{2}{*}{Decoding} & \multicolumn{3}{c}{Development} & \multicolumn{3}{c}{Evaluation-clean} & \multicolumn{3}{c}{Evaluation-other} \\ 
         & & \small CER & \small WER & \small sWER & \small CER & \small WER & \small sWER & \small CER & \small WER & \small sWER \\ \hline\hline
         \multicolumn{3}{l}{\textbf{Ksponspeech - phonetic}}\\
         \small Reproduction - Big Transf. & \small syllable & \small 6.471 & \small 16.760 & \small 11.929 & \small 7.779 & \small 20.965 & \small 12.874 & \small 8.260 & \small 24.354 & \small 14.408 \\ \hline
         \small K-Wav2vec 2.0 (Linear) & \small grapheme & \small 6.401 & \small 17.317 & \small 12.098 & \small 7.306 & \small 20.447 & \small 12.488 & \small 7.818 & \small 24.432 & \small 14.123 \\
         \small K-Wav2vec 2.0 (Linear) & \small syllable & \small 5.951 & \small 16.541 & \small 11.462 & \small 6.963 & \small 20.061 & \small 12.019 & \small 7.484 & \small 23.924 & \small 13.688 \\
         \small K-Wav2vec 2.0 (Transf.) & \small syllable & \small 5.976 & \small 16.556 & \small 11.512 & \small 6.985 & \small 20.081 & \small 12.024 & \small 7.536 & \small 24.057 & \small 13.848 \\
         \small K-Wav2vec 2.0 (Multi-task) & \small grapheme & \small 6.390 & \small 17.181 & \small 12.160 & \small 7.299 & \small 20.452 & \small 12.473 & \small 7.838 & \small 24.450 & \small 14.204 \\
         \small K-Wav2vec 2.0 (Multi-task) & \small syllable & \small 5.878 & \small 16.341 & \small 11.223 & \small 6.904 & \small 19.983 & \small 11.877 & \small 7.356 & \small 23.619 & \small 13.384 \\
         \small K-Wav2vec 2.0 (Multi-task) & \small joint & \small \textbf{5.867} & \small \textbf{16.248} & \small \textbf{11.157} & \small \textbf{6.881} & \small \textbf{19.895} & \small \textbf{11.765} & \small \textbf{7.339} & \small \textbf{23.527} & \small \textbf{13.273} \\ \hline\hline
         \multicolumn{3}{l}{\textbf{Ksponspeech - orthographic}}\\
         \small Reproduction - Big Transf. & \small syllable & \small 6.775 & \small \textbf{16.912} & \small 12.626 & \small 8.468 & \small 21.856 & \small 14.023 & \small 9.235 & \small 26.400 & \small 16.317 \\ \hline
         \small K-Wav2vec 2.0 (Linear) & \small grapheme & \small 6.776 & \small 17.954 & \small 12.964 & \small 7.893 & \small 21.528 & \small 13.632 & \small 8.904 & \small 26.517 & \small 16.351 \\
         \small K-Wav2vec 2.0 (Linear) & \small syllable & \small 6.326 & \small 17.455 & \small 12.364 & \small 7.498 & \textbf{20.954} & \small 13.039 & \small 8.431 & \small 25.807 & \small 15.648 \\
         \small K-Wav2vec 2.0 (Transf.) & \small syllable & \small \textbf{6.222} & \small 17.127 & \small \textbf{12.208} & \small \textbf{7.486} & \small 21.111 & \small 13.034 & \small 8.455 & \small 25.807 & \small 15.693 \\
         \small K-Wav2vec 2.0 (Multi-task) & \small grapheme & \small 6.926 & \small 18.320 & \small 13.292 & \small 8.052 & \small 21.832 & \small 13.961 & \small 9.095 & \small 27.039 & \small 16.753 \\
         \small K-Wav2vec 2.0 (Multi-task) & \small syllable & \small 6.261 & \small 17.166 & \small 12.274 & \small 7.534 & \small 20.983 & \small 13.093 & \small \textbf{8.366} & \small 25.701 & \small 15.427 \\
         \small K-Wav2vec 2.0 (Multi-task) & \small joint & \small 6.262 & \small 17.143 & \small 12.247 & \small 7.541 & \small 20.978 & \small \textbf{13.034} & \small 8.374 & \small \textbf{25.671} & \small \textbf{15.385} \\ 
    \bottomrule
    \end{tabular}{\parfillskip=0pt\par}
    \caption{Evaluation results on Ksponspeech development/evaluation-clean/evaluation-other sets with dual transcription. The K-Wav2vec 2.0 is a further pre-trained model on Ksponspeech training data. Models use either syllable or grapheme as a modeling unit except for the multi-task model. Considering inconsistent space in Korean transcription, we evaluated models with sWER as well as CER and WER.}
    \label{tab:high_resource}
\end{table*}

\subsection{High-resource evaluation}

We first evaluated the proposed model using the Ksponspeech dataset, which contains high-resource data with two different transcriptions. As a baseline, we reproduced a large transformer from a previous work \cite{bang2020ksponspeech} with the syllable modeling unit for both phonetic and orthographic transcripts. 
Table \ref{tab:high_resource} shows the performance of each model for the development, evaluation-clean, and evaluation-other datasets in terms of CER, WER, and sWER. Based on these results, we can derive the following observations. 

First, when other things are the same, the syllable-based models generally outperform the grapheme-based models for both transcripts. This is mainly because the grapheme-based models have more candidates to generate than the syllable-based models, and the quality of the decoded outputs can be degenerated, although a fairly large beam search is conducted. 

Second, the multi-task fine-tuning seemed to improve the Korean ASR performance even without the joint decoding strategy, and the performance improvement was the most noticeable with syllable decoding for phonetic transcripts. Because the model with hierarchical setting and multi-task learning captures the relation between syllable and grapheme information, syllable decoding of the multi-task model is complemented, especially in the transcript, where more repeated syllables and graphemes are provided without confusing terms.

Third, our proposed model, which is the last row in each transcript in the table, yielded the best performance for all evaluation metrics on all evaluation datasets when the phonetic transcript was used, where it showed the best performance on the evaluation-other dataset when orthographic transcripting was used. Although the proposed model is not the best for developing and evaluating a clean dataset, it still outperformed the benchmark model (i.e., the big transformer) in most cases. 

Notably, owing the orthographic transcript, because of numeric expressions and abbreviated notations in the orthographic transcript, all evaluation error rates were slightly higher than those on the phonetic transcript when other things were the same. These grammatical conversions in the script bring additional characters into the same utterance, confusing models to learn the distribution between speech and script. Without additional language models, our models learn grammatical patterns and outperform large transformers on evaluation sets in terms of all evaluation metrics.

Based on the aforementioned observations, we can conclude that, for high resource data, both components of our proposed model, namely the multi-task learning and joint decoding, not only contribute to the performance improvement individually, but also show a synergistic effect when they are used together.

\subsection{Low-resource evaluation}

\begin{table*}[t]
    \centering
    \begin{tabular}{lccccccc}
    \toprule 
         \multirow{2}{*}{Model} & \multirow{2}{*}{Decoding} & \multicolumn{3}{c}{Development} & \multicolumn{3}{c}{Evaluation} \\ 
         & & \small CER & \small WER & \small sWER & \small CER & \small WER & \small sWER \\ \hline\hline
         
         \textbf{Clovacall - Base} & & & & & & &  \\
         \small Paper - DS2\cite{ha2020clovacall} & \small syllable & & & & \small 11.4 & &  \\
         \small Paper - DS2 with SA\cite{ha2020clovacall} & \small syllable & & & & \small 10.1 & &  \\
         \small Paper - LAS \cite{ha2020clovacall} & \small syllable & & & & \small 15.1 & &  \\
         \small Paper - LAS with SA\cite{ha2020clovacall} & \small syllable & & & & \small 31.1 & &  \\
         \small Reproduction - Small Transf. & \small syllable & \small 1.024 & \small 3.817 & \small 1.619 & \small 11.237 & \small 26.862 & \small 19.817 \\ \hline
         \small K-Wav2vec 2.0 (Linear) & \small grapheme & \small 1.004 & \small 3.785 & \small 1.260 & \small 6.596 & \small \textbf{21.312} & \small 11.946 \\
         \small K-Wav2vec 2.0 (Linear) & \small syllable & \small 0.922 & \small 3.567 & \small 1.123 & \small 6.821 & \small 22.348 & \small 12.312 \\
         \small K-Wav2vec 2.0 (Transf.) & \small syllable & \small 0.918 & \small 3.462 & \small 1.073 & \small 13.439 & \small 29.987 & \small 24.461 \\
         \small K-Wav2vec 2.0 (Multi-task) & \small grapheme & \small 0.986 & \small 3.626 & \small 1.255 & \small 6.481 & \small 21.921 & \small 11.641 \\
         \small K-Wav2vec 2.0 (Multi-task) & \small syllable & \small \textbf{0.858} & \small \textbf{3.235} & \small \textbf{1.042} & \small 7.088 & \small 22.572 & \small 13.145 \\
         \small K-Wav2vec 2.0 (Multi-task) & \small joint & \small 0.875 & \small 3.285 & \small 1.055 & \small \textbf{6.414} & \small 21.475 & \small \textbf{11.499} \\

    \bottomrule
    \end{tabular}{\parfillskip=0pt\par}
    \caption{Evaluation results on Clovacall, where the amount of training data is limited. Previous results with or without SpecAugment(SA) are from \cite{ha2020clovacall}, and our baseline is a small transformer released by \cite{watanabe2018espnet}.} 
    \label{tab:low_resource}
\end{table*}

The second experiment was conducted on Clovacall, where there was insufficient training data to cover the vocabulary of the evaluation set. As a pre-trained model, we used the same pre-training strategy using Ksponspeech as an unlabeled speech data for language adaptation. Then, we fine-tuned the various configurations on Clovacall training data and evaluated the fine-tuned model on the test set in which the number of OOVs was large. As a baseline, we employed a small transformer with a beam search to decode syllable outputs. In previous studies \cite{ha2020clovacall}, DS2 \cite{amodei2016deep}, and LAS \cite{chan2016listen} models with SpecAugment \cite{park2019specaugment} were evaluated on Clovacall only in terms of CER.

Table \ref{tab:low_resource} shows that various configurations with further pre-trained models significantly outperform the models introduced in \cite{ha2020clovacall} and the baseline model. Because further pre-training with unlabeled data leads the model to better adapt to Korean, fine-tuning with small amounts of labeled data was found to be effective. The error gap between the development and evaluation set was much larger on Clovacall than on Ksponspeech. This large gap is not caused by the difference in the evaluation set but by the difference in the development set. The error rates on the evaluation set of Clovacall are similar to those on the evaluation-clean of Ksponspeech. The error rates on the development set of Clovacall, on the other hand, are much lower than those of the development set of Ksponspeech. There are two reasons for this enlarged gap between the development and evaluation datasets in a low-resource dataset. First, the size of the training dataset is insufficient, and the model does not have enough opportunities to train rare vocabularies. Although the probability of these rare vocabularies is also low in a high-resource dataset, the number of occurrences in the training dataset is larger in a high-resource dataset than in a low-resource one. Second, there are more OOVs in the evaluation dataset in a low-resource environment, which results in an inaccurate vocabulary prediction.

An interesting observation is that the best performance was reported with syllable decoding with multi-task learning for the development set, but the joint decoding with multi-task learning yielded higher error rates than the syllable decoding with multi-task learning. Because the OOV problem is not significant in the development set, only a syllable beam search without the help of a grapheme beam search is sufficient to predict the correct outputs. On the other hand, for the evaluation set, the syllable beam search was assisted by the grapheme beam search to generate the correct OOVs.

\subsection{Further pre-training evaluation}

\begin{table*}[t!]
    \centering
    \begin{tabular}{lcccccccccc}
    \toprule 
         \multirow{2}{*}{Model} & \multirow{2}{*}{Decoding} & \multicolumn{3}{c}{Development} & \multicolumn{3}{c}{Evaluation-clean} & \multicolumn{3}{c}{Evaluation-other} \\ 
         & & \small CER & \small WER & \small sWER & \small CER & \small WER & \small sWER & \small CER & \small WER & \small sWER \\ \hline\hline
         \textbf{Ksponspeech - phonetic} & & & & & & & & & & \\
         \small Reproduction - Big Transf. & \small syllable & \small 6.471 & \small 16.760 & \small 11.929 & \small 7.779 & \small 20.965 & \small 12.874 & \small 8.260 & \small 24.354 & \small 14.408 \\ \hline
         
         \small E-Wav2vec 2.0 (Transf.) & \small syllable & \small 7.512 & \small 20.433 & \small 14.639 & \small 8.588 & \small 24.033 & \small 15.341 & \small 9.236 & \small 28.197 & \small 17.617 \\
         \small E-Wav2vec 2.0 (Multi-task) & \small grapheme & \small 8.577 & \small 21.897 & \small 16.384 & \small 9.475 & \small 24.879 & \small 16.665 & \small 10.262 & \small 29.585 & \small 19.405 \\
         \small E-Wav2vec 2.0 (Multi-task) & \small syllable & \small 7.141 & \small 19.410 & \small 13.777 & \small 8.103 & \small 22.572 & \small 14.266 & \small 8.760 & \small 27.066 & \small 16.419 \\
         \small E-Wav2vec 2.0 (Multi-task) & \small joint & \small 7.146 & \small 19.402 & \small 13.726 & \small 8.100 & \small 22.480 & \small 14.203 & \small 8.812 & \small 27.103 & \small 16.538 \\ \hline
         
         \small S-Wav2vec 2.0 (Transf.) & \small syllable & \small 5.931 & \small 16.423 & \small 11.348 & \small 7.162 & \small 20.686 & \small 12.400 & \small 7.644 & \small 24.324 & \small 14.011 \\
         \small S-Wav2vec 2.0 (Multi-task) & \small grapheme & \small 6.770 & \small 17.966 & \small 12.949 & \small 7.688 & \small 21.214 & \small 13.260 & \small 8.445 & \small 25.849 & \small 15.592 \\
         \small S-Wav2vec 2.0 (Multi-task) & \small syllable & \small 6.099 & \small 16.775 & \small 11.731 & \small 7.118 & \small 20.325 & \small 12.327 & \small 7.715 & \small 24.439 & \small 14.238 \\
         \small S-Wav2vec 2.0 (Multi-task) & \small joint & \small 6.093 & \small 16.693 & \small 11.637 & \small 7.088 & \small 20.164 & \small 12.161 & \small 7.706 & \small 24.339 & \small 14.130 \\ \hline
         
         \small K-Wav2vec 2.0 (Transf.) & \small syllable & \small 5.976 & \small 16.556 & \small 11.512 & \small 6.985 & \small 20.081 & \small 12.024 & \small 7.536 & \small 24.057 & \small 13.848 \\
         \small K-Wav2vec 2.0 (Multi-task) & \small grapheme & \small 6.390 & \small 17.181 & \small 12.160 & \small 7.299 & \small 20.452 & \small 12.473 & \small 7.838 & \small 24.450 & \small 14.204 \\
         \small K-Wav2vec 2.0 (Multi-task) & \small syllable & \small 5.878 & \small 16.341 & \small 11.223 & \small 6.904 & \small 19.983 & \small 11.877 & \small 7.356 & \small 23.619 & \small 13.384 \\
         \small K-Wav2vec 2.0 (Multi-task) & \small joint & \small \textbf{5.867} & \small \textbf{16.248} & \small \textbf{11.157} & \small \textbf{6.881} & \small \textbf{19.895} & \small \textbf{11.765} & \small \textbf{7.339} & \small \textbf{23.527} & \small \textbf{13.273}  \\
         
    \bottomrule
    \end{tabular}{\parfillskip=0pt\par}
    \caption{Results of different pre-training methods with the Ksponspeech phonetic transcription. E-Wav2vec 2.0 is the English pre-trained model released by \cite{baevski2020wav2vec}. We pre-trained the base model on Ksponspeech training data from scratch, and we call this model S-Wav2vec 2.0. K-Wav2vec 2.0 is our proposed model, which continued pre-trained on the same amounts of unlabeled data as S-Wav2vec 2.0.}
    \label{tab:pre_training}
\end{table*}

We also compared our pre-training method, K-Wav2vec 2.0, to other pre-training approaches. In this experiment, we only used Ksponspeech training data to pre-train and fine-tune models by limiting the amount of both labeled and unlabeled data. We pre-trained the base architecture on 965 h of Ksponspeech training data from the initial state to obtain S-Wav2vec 2.0. We also compared our model to the English Wav2vec 2.0 (E-Wav2vec 2.0), which was pre-trained on Librispeech (960 h) released by \cite{baevski2020wav2vec}. We used the same setup as described in Section \ref{subsection:fine_tune_strategies} to fine-tune the models on labeled Ksponspeech data.

As shown in Table \ref{tab:pre_training}, our proposed pre-training models clearly outperformed the others. Moreover, adding the multi-task model with the joint decoder on the top of K-Wav2vec 2.0 improved performance and achieved the best overall performance on the Ksponspeech phonetic script. Compared with S-Wav2vec 2.0, further pre-training led to a significant improvement without requiring additional data by using pre-trained representations from another language. These results demonstrate that further pre-training is an effective approach for cross-lingual transfer through language adaptation.

\subsection{Appropriate data for pre-training}

\interfootnotelinepenalty=10000
\begin{table*}[t!]
    \centering
    \begin{tabular}{lllrrrc}
    \toprule 
         Data & Description & Style & \# Speaker & Amount & \# Audio & Len.(avg/min/max) \\ \hline
         Ksponspeech & \footnotesize Open-domain dialog corpus& \small Dialogue & 2000 & 965 hrs & 620000 & 5.60 / 0.99 / 30.99 \\
         Clovacall & \footnotesize Call-based speech for reservation & \small Dialogue & 11000 & 46 hrs & 59662 & 2.81 / 0.51 / 30.00 \\
         NIKL\tablefootnote{\url{https://corpus.korean.go.kr/main.do}} & \footnotesize Korean standard speech corpus& \small Reading & 118 & 150 hrs & 87045 & 6.23 / 0.48 / 50.98 \\
         Zeroth\tablefootnote{\url{https://www.openslr.org/40}} & \footnotesize Korean open-source speech corpus & \small Monologue & 181 & 51 hrs & 22263 & 8.35 / 3.30 / 20.66 \\
         ETRI\tablefootnote{\url{https://aiopen.etri.re.kr/service_dataset.php}} & \footnotesize Spontaneous speech dialogue & \small Dialogue & 200 & 37 hrs & 37200 & 3.62 / 2.09 / 16.57 \\
         VOTE400 & \footnotesize Reading speech of elderly people & \small Reading & 100 & 100 hrs & 111813 & 3.25 / 0.15 / 34.84 \\
         
    \bottomrule
    \end{tabular}{\parfillskip=0pt\par}
    \caption{Description of Korean speech datasets in public. Statistics are derived from the subset of data used for pre-training.}
    
    \label{tab:data_description}
\end{table*}

\begin{table*}[t!]
    \centering
    \begin{tabular}{lcccccccccc}
    \toprule 
         \multirow{2}{*}{Model} & \multirow{2}{*}{Decoding} & \multicolumn{3}{c}{Development} & \multicolumn{3}{c}{Evaluation-clean} & \multicolumn{3}{c}{Evaluation-other} \\ 
         & & \small CER & \small WER & \small sWER & \small CER & \small WER & \small sWER & \small CER & \small WER & \small sWER \\ \hline\hline
         \textbf{Ksponspeech - phonetic} & & & & & & & & & & \\
         \small Reproduction - Big Transf. & \small syllable & \small 6.471 & \small 16.760 & \small 11.929 & \small 7.779 & \small 20.965 & \small 12.874 & \small 8.260 & \small 24.354 & \small 14.408 \\ \hline
         
         \small 965H (Transf.) & \small syllable & \small 5.931 & \small 16.423 & \small 11.348 & \small 7.162 & \small 20.686 & \small 12.400 & \small 7.644 & \small 24.324 & \small 14.011 \\
         \small 965H (Multi-task) & \small grapheme & \small 6.770 & \small 17.966 & \small 12.949 & \small 7.688 & \small 21.214 & \small 13.260 & \small 8.445 & \small 25.849 & \small 15.592 \\
         \small 965H (Multi-task) & \small syllable & \small 6.099 & \small 16.775 & \small 11.731 & \small 7.118 & \small 20.325 & \small 12.327 & \small 7.715 & \small 24.439 & \small 14.238 \\
         \small 965H (Multi-task) & \small joint & \small 6.093 & \small 16.693 & \small 11.637 & \small 7.088 & \small 20.164 & \small 12.161 & \small 7.706 & \small 24.339 & \small 14.130 \\ \hline
         
         \small 1,249H (Transf.) & \small syllable & \small \textbf{5.809} & \small \textbf{16.209} & \small \textbf{11.208} & \small 6.875 & \small 20.071 & \small 11.828 & \small \textbf{7.306} & \small \textbf{23.449} & \small \textbf{13.277} \\
         \small 1,249H (Multi-task) & \small grapheme & \small 6.372 & \small 17.068 & \small 12.168 & \small 7.396 & \small 20.779 & \small 12.610 & \small 7.980 & \small 24.755 & \small 14.438 \\
        \small  1,249H (Multi-task) & \small syllable & \small 5.918 & \small 16.220 & \small 11.430 & \small 6.923 & \small 19.905 & \small 11.941 & \small 7.496 & \small 23.868 &   \small 13.599 \\
         \small 1,249H (Multi-task) & \small joint & \small 5.876 & \small 16.115 & \small 11.309 & \small \textbf{6.849} & \small \textbf{19.783} & \small \textbf{11.706} & \small 7.477 & \small 23.779 & \small 13.470 \\ \hline
         
         \small 1,349H (Transf.) & \small syllable & \small 5.921 & \small 16.341 & \small 11.290 & \small 7.063 & \small 20.100 & \small 11.950 & \small 7.453 & \small 24.098 & \small 13.703 \\
         \small 1,349H (Multi-task) & \small grapheme & \small 6.417 & \small 17.243 & \small 12.258 & \small 7.283 & \small 20.515 & \small 12.551 & \small 7.936 & \small 24.688 & \small 14.445 \\
         \small 1,349H (Multi-task) & \small syllable & \small 5.885 & \small 16.263 & \small 11.219 & \small 6.915 & \small 20.042 & \small 12.004 & \small 7.476 & \small 24.050 & \small 13.733 \\
         \small 1,349H (Multi-task) & \small joint & \small 5.893 & \small 16.260 & \small 11.231 & \small 6.934 & \small 19.871 & \small 11.897 & \small 7.442 & \small 23.924 & \small 13.599 \\
         
    \bottomrule
    \end{tabular}{\parfillskip=0pt\par}
    \caption{Results of different data combinations for pre-training. 965H is the pre-trained model with only Ksponspeech, and 1,349H indicates the pre-trained model with entire datasets including Ksponspeech, Clovacall, Zeroth, ETRI, NIKL, and VOTE400. 1,249H is the pre-trained model with our best combination, all datasets except VOTE400.}
    \label{tab:impact_of_data}
\end{table*}

To analyze the impact of the quality and quantity of training data on ASR performance, we investigated various combinations of Korean datasets for pre-training. In this experiment, we considered six publicly available datasets. Table \ref{tab:data_description} shows the statistics for each with a short description. These datasets have different topics, speakers’ ages, reading styles, and number of audio clips. The largest data is Ksponspeech \cite{bang2020ksponspeech}, as described in \ref{subsection:datasets}, consisting of open-domain dialog speeches recorded in a clean environment. ClovaCall \cite{ha2020clovacall}, a call-based dialog data for customer service, was also used for pre-training in this experiment. NIKL is a reading-style speech dataset released by the National Institute of Korean Language. It is intended to build a standard Korean speech corpus spoken by 120 people who lived in Seoul, the capital of Korea. Zeroth was developed by crowdsourcing, consisting of 51 h of speech data recorded using mobile phones. ETRI is a spontaneous speech dialog containing various conditions, such as a child’s voice, dialog in a driving car, and conversation in a noisy office. Unlike the other datasets, VOTE400 \cite{jang2021vote400} is a reading speech corpus of elderly people aged 65 or over, with slow and slurred pronunciation.

As a baseline, we used the model pre-trained on the 965 h of Ksponspeech training data, known as S-Wav2vec 2.0, as described in the previous section. We pre-trained the base architecture from scratch on all datasets apart from VOTE400, a total of 1,249 h of unlabeled data, to explore the benefit of additional data. The last model, pre-trained on 1,349 h of all datasets, was used to measure the impact of the unique dataset: VOTE400. All pre-trained models were fine-tuned on Ksponspeech with phonetic transcription, and stacks of transformers with a projection layer were used as a fine-tuning architecture of syllable output. We also used the multi-task fine-tuning architecture and evaluated the results of each unit separately, as well as the joint decoder.

Table \ref{tab:impact_of_data} shows that 1,249-h and 1,349-h pre-trained models generally outperformed 965-h models, taking advantage of more unlabeled data during pre-training. However, we observed that the 1,249-h multi-task model with the joint decoder achieved the best performance with 6.849/19.783/11.706 of CER/WER/sWER on the evaluation-clean set, respectively, instead of the 1,349-h models that used another 100 h of data for pre-training. Because VOTE400 has unique speech patterns of elderly people that differ greatly from other training data, adding irrelevant data during pre-training can disturb the performance of downstream tasks. This shows that adding more unlabeled data for pre-training is not the only factor affecting ASR performance, but the similarity between unlabeled data used in pre-training and the downstream task data in fine-tuning is an important factor.

\section{Discussions}
\label{sec:discussions}

\subsection{Intermediate relations in multi-task model}

\begin{figure}[t!]
\vskip 0.2in
\begin{center}
\centerline{\includegraphics[width=\columnwidth]{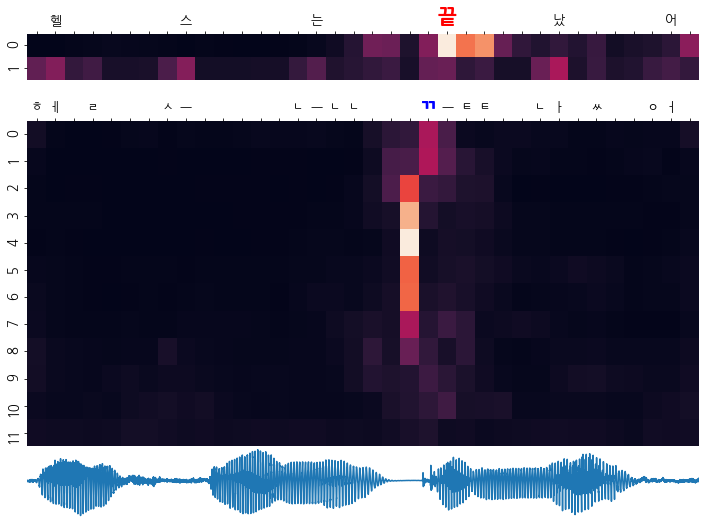}}
\caption{Visualization of attention maps generated from the multi-task model. In the middle of the figure, attentions are activated to relevant target grapheme highlighted blue. Attention maps of relational information between graphemes and target syllable are shown on the top.}
\label{fig:attention_map}
\end{center}
\vskip -0.2in
\end{figure}

To better understand intermediate representations between low-level and high-level units of the multi-task model, we visualized the attention maps of the transformer used in the base architecture (12 stacks of transformers) and the fine-tuning architecture (two transformer blocks). Figure \ref{fig:attention_map} shows the activated attention of an audio sample derived from Ksponspeech, where the multi-task model was trained on the Ksponspeech phonetic script. We placed attention maps from the base architecture and the corresponding graphemes in the middle of the figure, representing the relations between the audio and target grapheme highlighted in bold and blue colors. We observed that attention was activated in the region where it was pronounced. The other attention maps on the top of the figure were derived from the transformers of the multi-task architecture and represent the relation between graphemes and target syllable colored in red. Because the target grapheme corresponds to the first grapheme of the target syllable, attention is mostly activated around the target grapheme to gather relational information. This activation behavior indicates that the lower transformers of the multi-task model focus on grapheme-level information, and the higher transformers are activated to capture the intermediate relation. Because our proposed model learns the intermediate relations of multiple levels, representing a unique Korean writing structure, the multi-task model is able to generate multiple outputs without performance degradation.


\subsection{Analysis of contribution weight in joint decoder}

\begin{figure}[t!]
\vskip 0.2in
\begin{center}
\centerline{\includegraphics[width=\columnwidth]{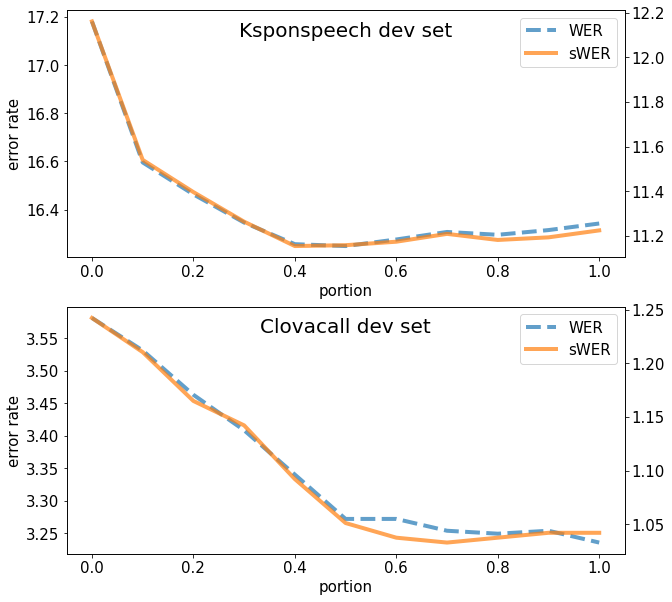}}
\caption{Error rates on the development set with different contribution weights($\gamma$) in joint decoding. The joint decoder with a contribution weight of 0 behaves the same as the grapheme decoder, and a contribution weight of 1 is the syllable decoder.}
\label{fig:portion_of_joint_decoder}
\end{center}
\vskip -0.2in
\end{figure}

To investigate the performance sensitivity with regard to the weight, $\gamma$, in the joint decoding, we conducted additional experiments by varying $\gamma$ from zero to one at an increment of 0.1 using joint decoding. We evaluated the multi-task model with the joint decoder on the Ksponspeech and Clovacall development set. 

Figure \ref{fig:portion_of_joint_decoder} shows that the joint decoding with $\gamma = 0.5$ yielded the best performance in terms of WER and sWER on the Ksponspeech. Additionally, the shape of performance figure is concave with regard to $\gamma = 0.5$, indicating that it is helpful to find the confident sequence among candidates by re-scoring them with the decoding results of different levels when the model is less confident in its predictions.

For Clovacall, on the other hand, it seems that the error rates monotonically decrease in accordance with the increase in $\gamma$. The best performance of WER was reported with $\gamma = 1.0$ (only with a syllable decoder), whereas that of sWER was reported with $\gamma = 0.7$, indicating that the role of the syllable decoder was more significant than that of the grapheme decoder. Because Clovacall had insufficient training data, the model overfitted and generated highly confident outputs on the development set sampled from the training set. Hence, the grapheme decoder did not have many opportunities to correct the possible errors made by the syllable decoder. 

Considering the various conditions in the real world, we picked a balanced $\gamma$ of 0.5 for other experiments because it not only yielded the best performance on the Ksponspeech, but it also achieved a reasonable performance on Clovacall. 

\subsection{Alleviating OOV problem with joint decoder}


\begin{table}[t!]
    \centering
    \begin{tabular}{lrrcc}
    \toprule 
         \multirow{2}{*}{} & \multirow{2}{*}{Total} & \multirow{2}{*}{OOV} & \multicolumn{2}{c}{Recovery} \\ 
         & & & Grapheme & Joint \\ \hline
         \# Vocab. & 542 & 33 & 5(15.1\%) & 5(15.1\%) \\ 
         \# Occur. & 20236 & 41 & 6(14.6\%) & 6(14.6\%) \\
          
    \bottomrule
    \end{tabular}{\parfillskip=0pt\par}
    \caption{Results of OOV syllables recovery on the Clovacall test set, varying the decoding strategy. OOV indicates out-of-vocabulary syllables possible to construct with grapheme vocabulary. 
    }
    \label{tab:oov_joint_decoding}
\end{table}

In this section, we investigate whether the joint decoder can help alleviate the OOV problem on the Clovacall test set, where apparent OOV syllables exist. We used the multi-task model to extract the beam search results of the graphene and joint decoder, respectively. For the joint decoder, we used $\gamma=0.5$ to utilize both the grapheme and syllable beam search results equally. Then, we evaluated each result on the test set and counted the number of correct OOV syllables. To focus on OOV restoration, we considered a total of 33 OOV syllables, which can be combined with our grapheme dictionary because the remaining seven syllables are impossible where corresponding graphemes are missing in the training set, as shown in Table \ref{tab:oov_information}.

Table \ref{tab:oov_joint_decoding} shows that both grapheme and joint decoding could alleviate the OOV problem by recovering 15.1\% of the OOV syllables with the grapheme combination. We also observed that the joint decoding restored 14.6\% of OOV occurrences as much as the grapheme decoding, where the total OOV occurrence was 41, and 6 were recovered on the test set. This shows that the beam search result of grapheme utilized in the joint decoding process alleviated the OOV problem when syllable decoding produced unconfident results for OOV syllables in speech. Because it is common to encounter the OOV problem in practice, the restoration property of the joint decoder can play an important role in a service-level ASR system.

\subsection{Cross-lingual transfer in further pre-training}
\label{subsection:cross_lingual_transfer_in_further_pre_training}

\begin{figure}[t!]
\vskip 0.2in
\begin{center}
\centerline{\includegraphics[width=\columnwidth]{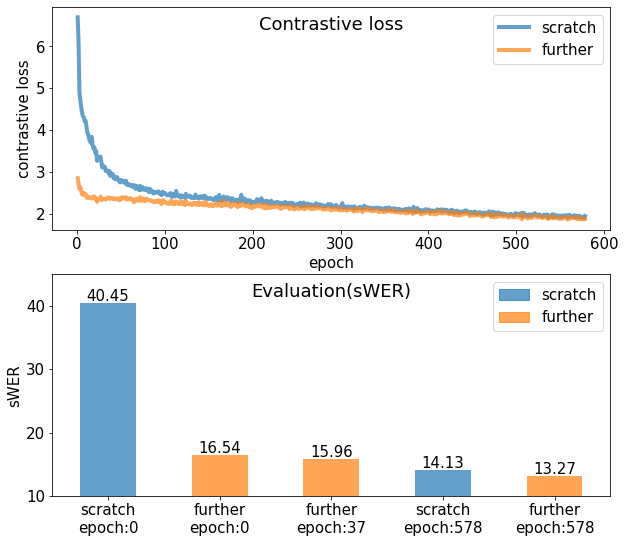}}
\caption{Contrastive loss and evaluation results with different pre-training methods. We recorded contrastive loss of validation set with different pre-training methods in every epoch. We evaluated the multi-task models of various pre-training checkpoints on the Ksponspeech evaluation-clean set.}
\label{fig:contrastive_loss_and_swer}
\end{center}
\vskip -0.2in
\end{figure}

In this section, we analyze the effectiveness of further pre-training in cross-lingual transfer in terms of contrastive loss and evaluation performance. We further pre-trained the English model on Ksponspeech training data and recorded the contrastive loss of the validation set in every epoch. We compared the pre-training approach to the other pre-training method, which pre-trained the base architecture from scratch. To analyze the pre-training results in a downstream task, we selected various checkpoints for each pre-training method and evaluated multi-task models with different checkpoints on the Ksponspeech evaluation-clean set. Figure \ref{fig:contrastive_loss_and_swer} shows the contrastive loss with different pre-training methods and the evaluation performance results of some checkpoints.

Because initialized states of the English models are useful, the additional pre-training method began with a lower validation loss than the from-scratch model and achieved 16.54 sWER on the downstream task. As the learning continues, the validation loss of the further pre-training method decreases, and the performance in ASR improves as the validation loss decreases. This indicates that further pre-training benefits downstream tasks by adapting the English model to Korean. 

At the end of the pre-training epoch, the further pre-training approach reached the lowest contrastive loss and achieved the best performance with a 13.27 sWER, whereas the pre-training method from scratch reached a slightly higher validation loss and achieved a 14.13 sWER. This shows that cross-lingual transfer with further pre-training was effective in performance improvement because it leveraged useful representations of the English model shared across languages.

\section{Conclusion}
\label{sec:conclusion}

In this work, we first presented a multi-task fine-tuning architecture with a joint decoder and a further pre-training approach for the Korean ASR system. Our experiments show that the multi-task model can generate multi-level outputs without performance degeneration, and our joint decoder enhances the ASR performance by overcoming the drawbacks of each modeling unit. On the large-scale speech data (Ksponspeech), our system achieved the best performance in terms of sWER on both phonetic and orthographic transcriptions. On the small-scale data (Clovacall), we also observed that our model with the joint decoder not only achieved the best and most robust performance, but it also alleviated the OOV problem. We also investigated cross-lingual transfer with a further pre-training approach, which was further pre-training the English model to Korean. This approach was effective by using pre-trained representations of the English model. The experimental results show that further pre-training outperforms the other pre-training approaches, wherein they use the same amount of data and GPU resources for gradient updates.

Despite some noticeable results, the current study has some limitations that lead us to consider future research directions. We conducted experiments to reduce contrastive loss by pre-training models to certain steps, owing to the limitations of time and GPU resources. However, as shown in Subsection \ref{subsection:cross_lingual_transfer_in_further_pre_training}, the contrastive loss continues to decrease as pre-training progresses, which appears to improve downstream performance. Therefore, future work should include evaluating the proposed architecture and decoding strategy with models further pre-trained until contrastive loss is no longer small. Additionally, future research should include the development of a decoding strategy that uses acoustic and linguistic information together, because the language model trained with various external data can further relieve the OOV problem by incorporating additional vocabularies.

\bibliography{main}
\bibliographystyle{icml2021}
\clearpage

\end{document}